\documentclass[10pt,twocolumn,letterpaper]{article}

\usepackage[pagenumbers]{cvpr} %

\usepackage[dvipsnames]{xcolor}

\newif\ifdraft
\drafttrue

\ifdraft

\newcommand{\kac}[1]{{\color{orange}[\textbf{Kfir:} 
\textit{#1}]}}

\else
\newcommand{\pcc}[1]{}
\newcommand{\kac}[1]{}

\fi

\usepackage{multirow}
\usepackage{amsmath}
\usepackage{amsthm}
\usepackage[makeroom]{cancel}

\newtheorem{theorem}{Theorem}[section]

\newtheorem{definition}[theorem]{Definition}

\usepackage{makecell}
\usepackage{indentfirst}
\usepackage[dvipsnames]{xcolor}

\usepackage{booktabs}
\usepackage{multirow}
\usepackage{makecell}
\usepackage{geometry}
\usepackage{caption}
\usepackage{booktabs,arydshln}
\usepackage{amssymb}%
\usepackage{pifont}%

\makeatletter
\def\adl@drawiv#1#2#3{%
        \hskip.5\tabcolsep
        \xleaders#3{#2.5\@tempdimb #1{1}#2.5\@tempdimb}%
                #2\z@ plus1fil minus1fil\relax
        \hskip.5\tabcolsep}
\newcommand{\cdashlinelr}[1]{%
  \noalign{\vskip\aboverulesep
           \global\let\@dashdrawstore\adl@draw
           \global\let\adl@draw\adl@drawiv}
  \cdashline{#1}
  \noalign{\global\let\adl@draw\@dashdrawstore
           \vskip\belowrulesep}}
\makeatother

\definecolor{cvprblue}{rgb}{0.21,0.49,0.74}
\usepackage[pagebackref,breaklinks,colorlinks,allcolors=cvprblue]{hyperref}
\usepackage{tcolorbox}
\tcbuselibrary{skins, breakable}
\usepackage{graphicx}
\usepackage{algorithm}
\usepackage{algpseudocode}
\usepackage{multirow}
\usepackage{titletoc}
\usepackage{minitoc}

\def\paperID{8812} %
\def\confName{CVPR}
\def\confYear{2026}

\title{Diverse Video Generation with Determinantal Point Process-Guided Policy Optimization}
\author{
Tahira Kazimi \hspace{1mm}
Connor Dunlop \hspace{1mm}
Pinar Yanardag \\
Virginia Tech \\
{\tt\small \{tahirakazimi, cdunlop, pinary\}@vt.edu}
\\
{\href{https://diverse-video.github.io/}{\small \texttt{diverse-video.github.io}}}
}

\begin{document}

\twocolumn[{
\maketitle
\begin{center}
    \captionsetup{type=figure}
    \vspace{-1em}
\newcommand{\imwidth}{1\textwidth}

\begin{tabular}{@{}c@{}}
 
\parbox{\imwidth}{\includegraphics[width=\imwidth, ]{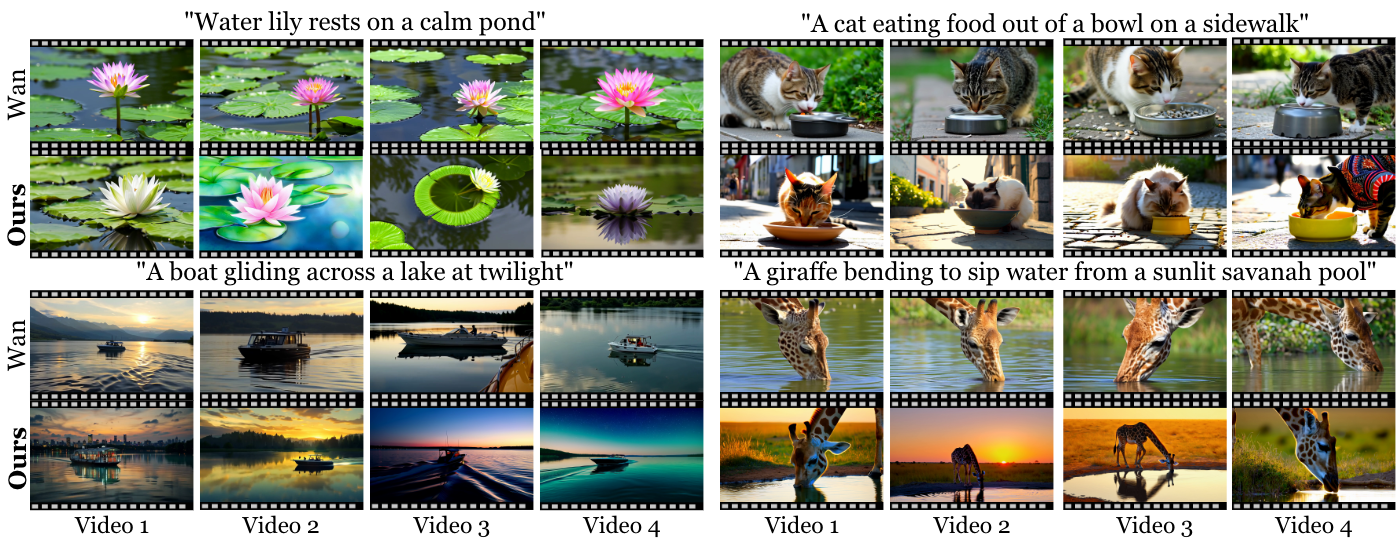}}
\\

\vspace{1em}
\end{tabular}
    \vspace{-3em}
    \captionof{figure}{Given an input prompt, \texttt{DPP-GRPO} enables generation of diverse sets of videos spanning cinematic factors such as \textit{camera motion} or \textit{scene layout}. We formulate diversity as set-level policy optimization with a DPP-based diminishing-returns term to suppress similar videos as the set grows. For brevity, we display only the first frame of each clip; see the supplement for frame-by-frame comparisons and videos. \textit{We note that our results are uncurated and shown in the exact order produced by our method.}} 
 
    \label{fig:teaser}
    \vspace{-0.5em}
\end{center}
}]

\maketitle
\begin{abstract}
While recent text-to-video (T2V) diffusion models  have achieved impressive quality and prompt alignment, they often produce low-diversity outputs when sampling multiple videos from a single text prompt. We tackle this challenge by formulating it as a set-level policy optimization problem, with the goal of training a policy that can cover the diverse range of plausible outcomes for a given prompt. To address this, we introduce \textit{DPP-GRPO}, a novel   framework for diverse video generation that combines Determinantal Point Processes (DPPs) and Group Relative Policy Optimization (GRPO) theories to enforce explicit reward on diverse generations.  Our objective turns diversity into an explicit signal by imposing diminishing returns on redundant samples (via DPP) while supplies groupwise feedback over candidate sets (via GRPO). Our framework is plug-and-play and model-agnostic, and  encourages diverse generations across visual appearance,  camera motions, and scene structure without sacrificing prompt fidelity or perceptual quality. We implement our method on WAN and CogVideoX, and show that our method consistently improves video diversity on state-of-the-art benchmarks such as VBench, VideoScore, and human preference studies. Moreover, we release our code and a new benchmark dataset of 30,000 diverse prompts to support future research.

    \vspace{-2em}

\end{abstract}
 
\section{Introduction}

Text-to-video  diffusion models have rapidly improved, enabling the generation of visually coherent and high-fidelity short videos across a range of applications including entertainment, education, advertising, and social media content creation~\cite{melnik2024video, wan2025wan, yang2024cogvideox, DeepMindVeo3TechReport2025}.   However, despite these advances, current video generators often suffer from limited diversity in their outputs. They tend to produce videos that conform to a narrow distribution of styles or scenarios. For instance, although a model may generate accurate scenes of `\textit{A giraffe bending to sip water from a sunlit Savannah pool}' (see Fig. \ref{fig:teaser}), it repeatedly favors similar types of scenes and motions, overlooking many other plausible variations.

The lack of diversity in generative outputs has been a long-standing issue, especially in image synthesis research where output distributions collapse toward a few modes  \cite{thanh2020catastrophic, kossale2022mode}. Prior efforts tackled diversity with strategies such as entropy-based sampling~\cite{jalali2025sparke}, dataset-coverage objectives~\cite{Dombrowski_2025_CVPR}, group sampling~\cite{parmar2025scaling}, or noise-injection schemes~\cite{sadat2023cads}. However, these techniques are largely tailored to static imagery and face key limitations in video: many rely on expensive test-time optimization and memory-intensive caches of prior latents ~\cite{jalali2025sparke} which is ill-suited for video. On the other hand, methods that require access to the full training set~\cite{Dombrowski_2025_CVPR} or mandating architectural changes~\cite{parmar2025scaling} are infeasible due to significant computational overhead.  Moreover, these methods overlook challenges unique to video. Beyond visual appearance, generated videos must vary along temporal and cinematic dimensions: such as  \textit{object motion, camera movement, scene structure}. As a result, users often rely on prompt engineering, and exhaustive sweeps of seeds/guidance settings to find diverse generations. While these strategies can surface occasional diversity, they require substantial time and compute costs with inconsistent gains. 

How can we generate a set of outputs that varies factors such as \textit{motion, scene composition}, and \textit{camera language} while remaining  faithful to the input prompt? In this paper, we approach this problem as a set-level policy optimization task and introduce \texttt{DPP-GRPO}, a diversity-aware framework that couples a DPP ~\cite{Macchi_1975, kulesza2012determinantal} with GRPO \cite{shao2024deepseekmath} objective. Our DPP component injects diminishing returns, explicitly discounting similar samples so the first instance of a choice (e.g., \textit{dolly shot}) is rewarded while redundant variants add diminishing returns. Then GRPO supplies groupwise feedback over a batch of candidates to prefer sets  that jointly span semantically diverse dimensions in user input.  Our method is model-agnostic and plug-and-play: it can be used in open-source  (e.g., WAN, CogVideoX) or black-box models (e.g., Veo) to enable generation of faithful, high-quality videos  with significantly improved diversity. Our contributions are as follows.

\begin{itemize} 
    \item We formulate diverse video generation as a set-level optimization problem and introduce a novel policy optimization framework that combines GRPO's group-wise optimization with a DPP-based diminishing-returns diversity objective, enabling a policy that generates semantically faithful yet non-redundant prompt sets. To our knowledge, this is the first work tackling diversity problem in video generation.
    
    \item Our method is model agnostic and plug-and-play, and requires no architectural changes and applies to open models (Wan, CogVideoX) and to black-box APIs (e.g., Veo).
    \item Across standard T2V benchmarks and human studies, we show consistent improvements in set diversity  while preserving fidelity and temporal coherence.
    \item We release a curated dataset of 30K diverse prompt–variant pairs designed specifically for diverse video generation, providing the first benchmark resource for this emerging problem.
\end{itemize}

\begin{figure}%
\centering
\includegraphics[width=.476\textwidth]{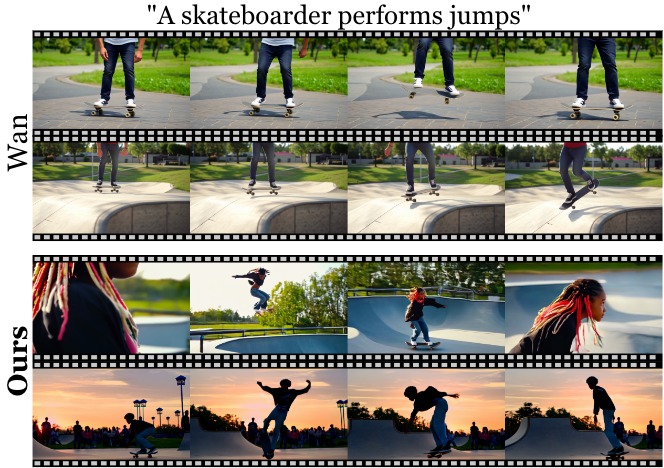}
    \vspace{-0.75em} 
    \caption{Frame-level examples illustrating motion, subject, and camera diversity achieved by \texttt{DPP-GRPO}. Given the same input prompt, our method generates videos that vary in subject \textit{appearance, environment}, and \textit{camera movement} while maintaining semantic consistency with the original prompt.
    }\vspace{-1em}
\label{fig:motion}
\end{figure}

\section{Related Work}

\noindent \textbf{Diversity in Generative Models} Balancing fidelity and diversity is an important challenge in generative modeling. In diffusion models, guidance mechanisms like classifier-free guidance (CFG)~\cite{sadat2024no, ho2022classifier} improve prompt alignment but reduce sampling stochasticity, often suppressing diversity~\cite{jalali2025sparke}.
In the image domain, several strategies have been proposed; ~\cite{sehwag2022generating} sample from low-density regions to encourage diversity, but their method operates in pixel space and does not transfer to latent diffusion. Reinforcement learning–based methods~\cite{miao2024training} optimize for diversity via reward signals, yet require full retraining. Several other works ~\cite{jalali2025sparke, askari2024improving, jalali2024conditional} incorporate diversity-aware objectives into generation but depend on internal model access, limiting scalability and deployment. DreamDistribution~\cite{zhao2023dreamdistribution} promotes variation by learning composite prompt embeddings from few-shot examples, while ~\cite{Dombrowski_2025_CVPR} formulate it as a dataset coverage problem requiring training data access. \textit{To the best of our knowledge, no prior work directly targets diversity in video generation, a setting that introduces challenges unique to the video domain.}  %

\noindent \textbf{RL-based Video Generation} 
Prior work adapts preference learning to video along several axes.  VideoDPO~\cite{liu2025videodpo} extends DPO to text-to-video diffusion with an omni-preference objective balancing visual quality and text relevance. Flow-DPO \cite{liu2025improving} aligns flow-matching video generators using a multi-dimensional reward,  improving motion smoothness and prompt adherence over SFT and alternative RLHF variants. More fine-grained supervision has also been explored via DenseDPO \cite{wu2025densedpo}, which targets temporal locality by densifying video preferences across time. Complementary directions include  GRPO variants tailored to video control, such as  DanceGRPO \cite{xue2025dancegrpo} for learning temporally coherent, motion-centric policies from groupwise relative feedback. Prompt-a-Video \cite{ji2025prompt} explores prompt-level control and conditioning strategies for text-to-video models. \textit{However, none of these methods  address diversity of generated outputs in videos which is the main focus of our work.}

\section{Background}
\label{sec:background}
\begin{figure}%
\centering
\includegraphics[width=0.50\textwidth]{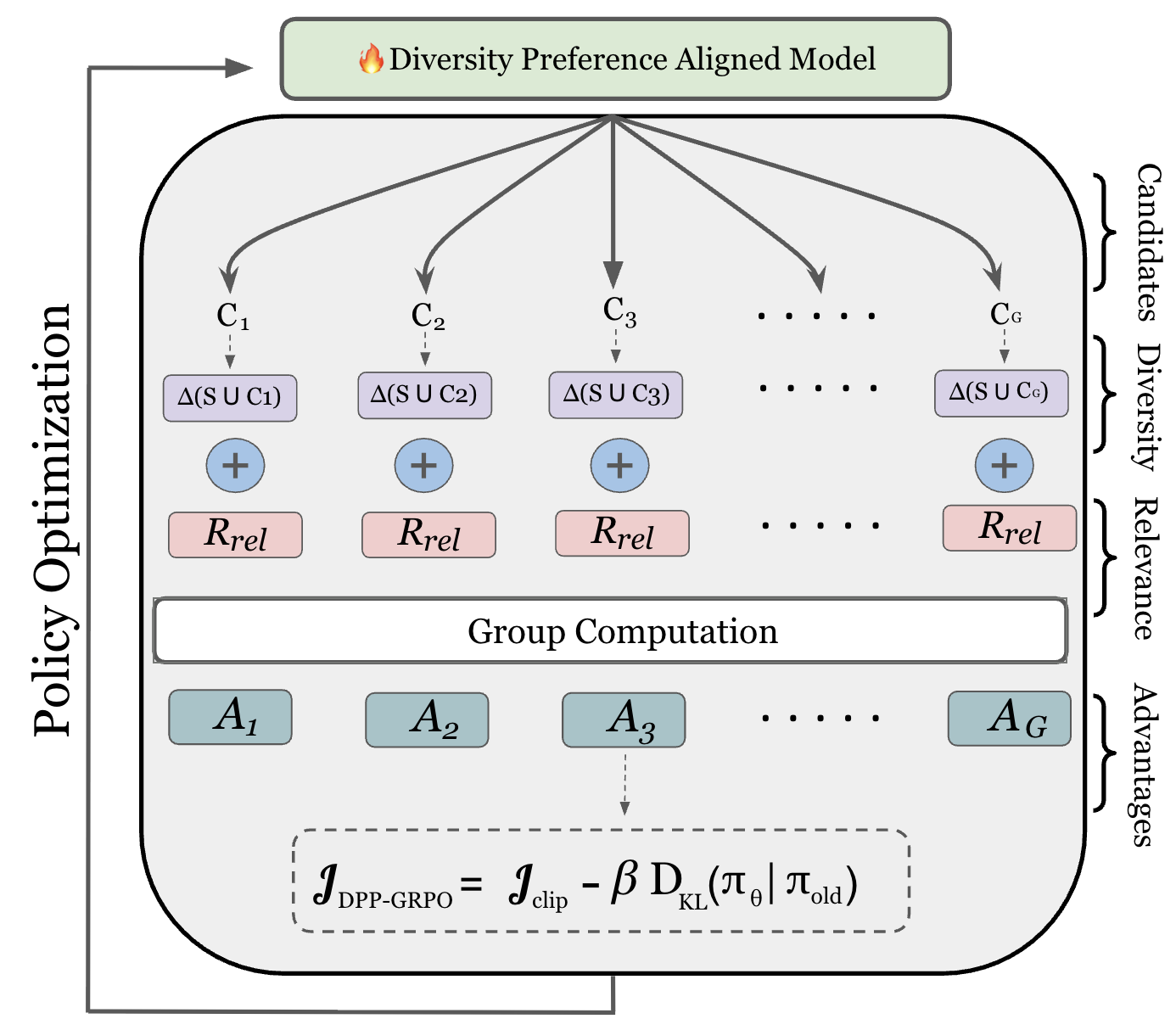}
    \vspace{-1em}
    \caption{\textbf{Framework Overview.} The model generates a group of $G$ candidates $(C_1,\ldots,C_G)$ for each prompt. 
Each candidate is scored by a composite reward combining DPP marginal gain $\Delta(S \cup C_i)$ 
and relevance $R_{\text{rel}}(C_i)$. Groupwise normalization produces advantages $(A_i)$, which 
update the policy under the DPP--GRPO objective.
    }\vspace{-1em}
\label{fig:framework}
\end{figure}

\paragraph{Determinantal Point Processes}
We quantify the diversity of a set of generated responses using concepts DPPs, which are well-suited for modeling diversity and negative correlation in subsets \cite{kulesza2012determinantal}. In particular, we focus on the L-ensemble formulation~\cite{borodin2005eynard,kulesza2012determinantal}, a subclass of DPPs widely used in machine learning.
\begin{definition}
\label{def:dpp}
Let $\mathcal{Y} = \{1, 2, \ldots, N\}$ denote a finite ground set, and let $\mathbf{Y} \subseteq \mathcal{Y}$ be a random subset. Suppose $L \in \mathbb{R}^{N \times N}$ is a real symmetric positive semi-definite matrix. Then $L$ defines a DPP in the form of an L-ensemble if, for any subset $S \subseteq \mathcal{Y}$,
\[
\Pr(\mathbf{Y} = S) \propto \det(L_S),
\]
where $L_S$ is the principal submatrix of $L$ indexed by elements in $S$. 
\end{definition}
If the entries of the matrix $L$ are interpreted as similarity scores (e.g., cosine similarity) between elements in the ground set, then the determinant $\det(L_S)$ can be viewed as the squared volume spanned by the feature vectors of the elements in subset $S$. This volume increases when the vectors are diverse (i.e., linearly independent) and shrinks when they are similar or redundant. Hence, DPPs naturally favor subsets with high internal diversity.
\vspace{-1.5em}

\paragraph{Group Relative Policy Optimization}
GRPO~\cite{shao2024deepseekmath} is a value-free reinforcement learning framework that computes advantages through group-based normalization of sampled outputs' rewards, eliminating the need for a separate critic network. Given a query $q$, the method samples $G$ responses $p_1, \ldots, p_G$ from the old policy $\pi_{\theta_{\text{old}}}$ and updates the current policy $\pi_\theta$ to maximize:
\begin{equation}
\begin{split}
\mathcal{J}_{\text{GRPO}}(\theta) = \mathbb{E}_{a, \{o_i\}} \Bigg[ \frac{1}{G} \sum_{i=1}^{G} \min \left( \frac{\pi_\theta(o_i|q)}{\pi_{\theta_{\text{old}}}(o_i|q)} A_i, \right. \\
\left. \text{clip}\left( \frac{\pi_\theta(o_i|q)}{\pi_{\theta_{\text{old}}}(o_i|q)}, 1\!-\!\epsilon, 1\!+\!\epsilon \right) A_i \right) - \beta \mathbb{D}_{\text{KL}}(\pi_\theta \| \pi_{\text{ref}}) \Bigg]
\end{split}
\label{eq:grpo}
\end{equation}

Advantages $A_i$ are computed by normalizing rewards $r_i$ within the group:
\begin{equation}
A_i = \frac{r_i - \operatorname{mean}(r_{1:G})}{\operatorname{std}(r_{1:G})}
\end{equation}
Standard GRPO (Eq.~\ref{eq:grpo}) objective optimizes policies to maximize expected reward, with advantage normalization encouraging high probability on sample with maximum reward. While effective for quality optimization, this can lead to diversity collapse where policies consistently generate similar high-reward outputs rather than exploring diverse valid responses.

\section{Methodology}
\label{sec:method}
Given a user prompt $q$, a T2V model $\mathcal{G}$, and a desired output size $K$, our goal is to generate a set of prompts $\mathcal{P}_q = \{p_1, \ldots, p_K\}$ whose corresponding videos $\mathcal{V}_q = \{\mathcal{G}(p_i)\}_{i=1}^{K}$ are diverse yet faithful to the intent of $q$. We approach this through set-level policy optimization, where each generated response is evaluated based on the additional diversity it contributes to the partially constructed set, encouraging exploration of complementary variations while avoiding redundancy.
To accomplish this, we introduce \texttt{DPP-GRPO}, a model-agnostic alignment framework that trains a policy using a diversity-aware DPP \cite{Macchi_1975, kulesza2012determinantal} reward in combination with the GRPO objective \cite{shao2024deepseekmath}. Our DPP component enforces diminishing-returns property: the first appearance of a novel factor (e.g., a dolly shot) receives higher reward, while subsequent redundant variants yield progressively smaller gains. Then, GRPO supplies group-wise relative feedback over a batch of candidates, guiding the policy toward sets that jointly span diverse dimensions such as \textit{camera motion} or \textit{scene layout. }

While our method could be applied inside a T2V generator such as by training the model to  diversify outputs according to our objective, we instead optimize a prompt-based policy for three practical reasons: (i) \textit{efficiency} since we do not perform backprop through video sampling, which yields significantly shorter training time, (ii) \textit{cinematic expressivity} since factors such as camera motion, scene composition are naturally controllable via prompt tokens, and (iii) \textit{plug-and-play deployment} since the prompts generated by our method  can work with open-source and black-box models without architectural changes or access to latents/gradients.

\subsection{DPP-GRPO}
\label{sec:reward}

\texttt{DPP-GRPO} balances two competing objectives: (1) maximizing diversity, and (2) maintaining semantic alignment with the user's input prompt.  We formulate diverse video generation as a \textit{set-level} optimization problem: given a target set size $K$, our goal is to generate a set of $K$ video prompts that captures a wide range of cinematic variations including \textit{motion, composition}, and \textit{perspective} which can then be used with any T2V generator. To quantify and promote diversity within this set, we incorporate DPPs, which has been shown to be more effective than simple pairwise distance metrics in modeling diverse subsets \cite{parker2020effective}. Following  Def. \ref{def:dpp}, we define diversity of a set $p_{1:k}$ as:
\begin{equation}
\mathrm{Div}(p_{1:k}) = \log \det(L_\phi(p_{1:k}) + I),
\label{eq:div_DPP}
\end{equation}
where $L_\phi[p_i, p_j] = f(\phi(p_i), \phi(p_j))$, $f$ is a kernel function and $\phi(.)$ is a selected embedding model \cite{reimers2019sentence} which can map a response into a high-dimensional semantic space. We define $f$ to be normalized cosine similarity and regularize $L_\phi + I$, to avoid singular matrices, following \cite{kulesza2012determinantal}. The log-determinant measures the log-volume spanned by $\{\phi(p_1), \ldots, \phi(p_k)\}$, increasing when prompts are diverse and decreasing when overly similar. According to GRPO training, we sample $G$ responses per query and assign each $p_i$ a marginal diversity gain relative to a reference set $\mathcal{R}_q$:
\begin{equation}
\Delta(p_i \mid \mathcal{R}_q) = \mathrm{Div}(\mathcal{R}_q \cup \{p_i\}) - \mathrm{Div}(\mathcal{R}_q)
\label{eq:marginal_div}
\end{equation}
Formally, a response $p_i$ has a higher marginal gain if $p_i$ spans a dimension not covered by elements in $\mathcal{R}_q$, increasing the volume spanned by existing set members. 
Here, $\mathcal{R}_q$ consists of curated ground-truth variants for $q$ that capture distinct semantic and cinematic modes (e.g., subject changes, camera motions, or scene layouts). Each pair $e_i, e_j \in \mathcal{R}_q$ is both semantically aligned with the query and mutually diverse, so optimizing marginal gains relative to $\mathcal{R}_q$ teaches the policy to distribute candidates across these modes rather than collapsing onto a single one.
As a result, the final diversity-aware reward signal takes the final closed-form,
\begin{equation}
\Delta(p_i \mid \mathcal{R}_q) = \log\det(L_\phi(\mathcal{R}_q \cup \{p_i\})) - \log\det(L_\phi(\mathcal{R}_q))
\label{eq:marginal_div_final}
\end{equation}
which is used by GRPO objective \textit{($+ I$ term is omitted for simplicity in notation)}.

\paragraph{Relevance.}
To ensure that generated prompts remain semantically faithful to the user's intent, we define a relevance term in our reward mechanism that encourages alignment with both the user-provided query $q$ and elements in $\mathcal{R}_q$. For instance, given the input ``\textit{a dog is playing with a ball at the beach},'' a high-quality variant like ``\textit{a dolly shot of a poodle is playing with a red ball at the beach}'' should receive a high relevance score, while trivial copies (e.g., ``\textit{a dog is playing near the beach with a ball}'') or unrelated variants should be penalized. We compute this via considering two similarities: one between the generated prompt $p_i$ and the user query $q$, and another between $p_i$ and elements in $\mathcal{R}_q$:
\begin{equation}
R_\text{rel}=\frac{1}{|\mathcal{R}_q|} \sum_{g \in \mathcal{R}_q}\cos(\phi(p_i), \phi(q)) \cdot   \cos(\phi(p_i), \phi(g))
\label{eq:relevance_avg}
\end{equation}
This  formulation enforces a joint constraint: $p_i$ must maintain high similarity to both the original query and valid variations. %
Our final objective is then a composite reward mechanism in training that balances diversity and semantic alignment:
\begin{equation}
R(p \mid q, g) = \lambda_{\text{div}} \Delta(p_i \mid \mathcal{R}_q) + \lambda_{\text{rel}} R_{\text{rel}}
\label{eq:combined_reward}
\end{equation}
where $\lambda_{\text{div}}$ and $\lambda_{\text{rel}}$ are weighting coefficients. Each component targets a specific failure mode: $\Delta$ prevents mode collapse through diversity enforcement, while $R_{\text{rel}}$ maintains semantic fidelity to the original query and ground-truth samples.

\noindent \textbf{Training.}
We follow a two-stage post-training  by applying supervised finetuning to initialize the policy model before reinforcement learning. The finetuning corpus $\mathcal{D} = \{(x,y)\}$ consists of user prompts paired with diverse variants generated via chain-of-thought prompting \cite{wei2022chain} (see details below). Finetuning is performed for up to 50 iterations. 
After a warm start, following GRPO~\cite{shao2024deepseekmath}, given a user prompt $q$, we sample $G$ responses $p_1, \ldots, p_G$ from the old policy $\pi_{\theta_{\text{old}}}$ and update the current policy $\pi_\theta$ to maximize the GRPO objective (Eq.~\ref{eq:grpo}). Each training sample $D_i = (s, q_i, \mathcal{R}_q)$ contains a system prompt $s$, a user query $q_i$, and the reference set $\mathcal{R}_q$. Rewards for sampled outputs $p_1, \ldots, p_G$ are computed using Eq.~\ref{eq:combined_reward} and used to update $\pi_\theta$ via Eq.~\ref{eq:grpo}. Through this process, the policy learns a transferable rule: how to propose a new candidate that expands the semantic coverage of an existing reference set while remaining faithful to the original query. \textit{An ablation on reference set $\mathcal{R}_q$ is provided in SM.}

\noindent \textbf{Inference.}
Given an input prompt $q$ and target output size $K$, we initialize an empty reference set $\mathcal{R}_q = \emptyset$. The first sample $p_1$ is generated conditioned only on $(s, q)$ and added to the reference set. In subsequent steps, we iteratively provide $(s, q, \mathcal{R}_q)$ to generate $p_2, p_3, \ldots$, expanding $\mathcal{R}_q$ with each new variant until $K$ prompts are produced. This autoregressive rollout mirrors the structure used during training: each new sample is evaluated relative to the partially constructed set, allowing the learned diversification strategy to operate without requiring access to the curated reference sets used during training.

\noindent \textbf{Dataset.} We curate a diverse video-prompt dataset using GPT-5-nano~\cite{achiam2023gpt}. Data construction proceeds in two stages: (1) we generate 3K base prompts using GPT-5-nano, and (2) for each base prompt, rather than relying on simple expansion prompts, we design an agentic pipeline that iteratively refines diversity. An architect agent proposes diverse variations, while a critic agent evaluates each candidate using video-level diversity and quality metrics (TIE, TCE, CLIP), ensuring that resulting prompts are grounded in video behavior rather than text alone. We synthesize 10 semantically diverse variations per base prompt, yielding 30K samples in total.

\section{Experiments}

We showcase the effectiveness of \texttt{DPP-GRPO} on two state-of-the-art T2V models Wan2.1~\cite{wan2025wan} and CogVideoX~\cite{yang2024cogvideox}, as well as black-box T2V model VEO3~\cite{DeepMindVeo3TechReport2025}\footnote{Since VEO3 is prohibitively expensive (on fal.ai, it costs ~\$3,000 to produce 1,000 videos required by our quantitative analysis), our evaluation on Veo3 is restricted to qualitative comparisons.}. Since no existing method directly addresses diverse video set generation, we benchmark against RL-driven prompt optimization baselines    Promptist~\cite{hao2023optimizing}, Prompt-A-Video~\cite{ji2024prompt}) and GPT-5.

\begin{figure*}[!htb]
\centering
\includegraphics[width=\textwidth]{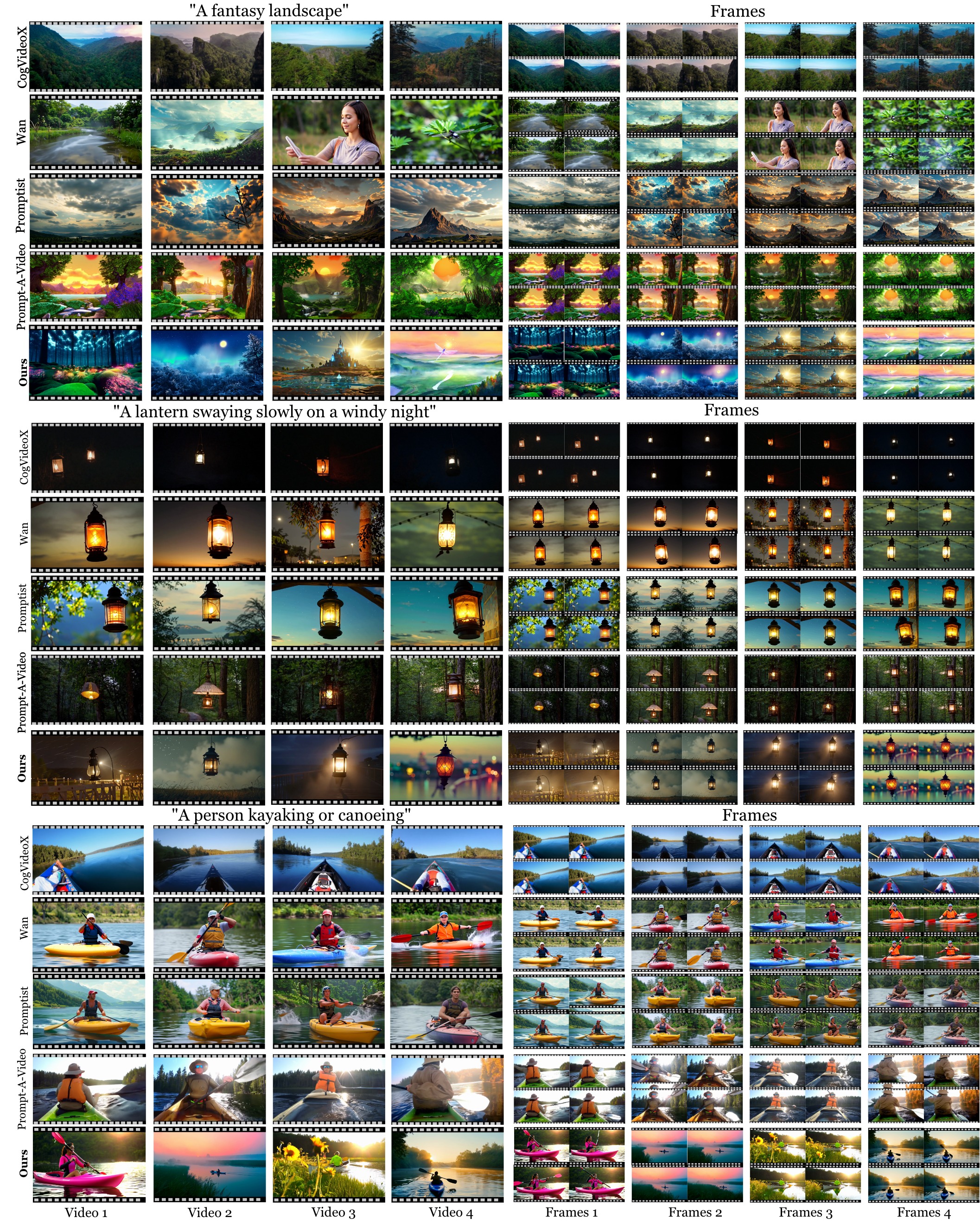}
    \vspace{-1em}
    \caption{\textbf{Qualitative Comparison.} \texttt{DPP-GRPO} diversifies several cinematic factors such as subject, scene, motion, and camera-view diversity while preserving the prompt alignment and achieves more diverse and semantically faithful videos compared to baselines. (a) For clarity, we provide the first frames of each video (b) Detailed frame-by-frame comparisons of the same videos are given (\textit{please kindly zoom-in for details}). Please visit our SM for more comparisons and high-quality videos.}
    \vspace{-1em}
\label{fig:qual}
\end{figure*}

\noindent \textbf{Experimental Setup} All experiments are conducted on four NVIDIA L40S GPUs. We first perform supervised fine-tuning for 50 iterations, followed by GRPO training for approximately 1,200 iterations. The learning rate is set to $2\times10^{-5}$ for supervised fine-tuning and $2\times10^{-7}$ during GRPO optimization, with $\lambda_{\text{div}} = \lambda_{\text{rel}} = 0.5$ by default.
Our dataset comprises 30K user query–sample pairs designed to capture diverse generative outcomes.
For video generation, we utilize Wan2.1 \cite{wan2025wan} and CogVideoX \cite{yang2024cogvideox} as our backbone models, and Qwen2-7b-Instruct~\cite{yang2025qwen3} as our alignment model. Unless otherwise specified, we adopt a guidance scale of 6, with 40 inference steps and 81 output frames per video.

\noindent \textbf{Qualitative Experiments}
We present qualitative results across a range of prompts, comparing \texttt{DPP-GRPO} to video baselines, CogVideoX~\cite{yang2024cogvideox} and Wan~\cite{wan2025wan}. Because our method optimizes set-level diversity with a DPP-based diminishing-returns objective, it directly \textit{produces diverse sets without post-hoc cherry-picking}. All examples are shown \textit{uncurated and in the exact generation order}, demonstrating that our outputs are diverse by design.

Fig.\ref{fig:teaser} shows a comparison between our method and Wan model, where our method consistently yields sets with substantial variation in factors such as scene composition while preserving semantic fidelity to the prompt (\textit{please refer to SM for frame-by-frame comparisons, high-quality videos, and prompts associated with each sample.}).  Moreover, we provide qualitative comparisons across multiple models, including Wan, CogVideoX, Promptist, and Prompt-A-Video (see Fig. \ref{fig:qual}).   We highlight two failure modes that other methods suffer:  \textit{homogeneous output generation} and \textit{weak semantic grounding}. For instance, in the abstract prompt \textit{``a fantasy landscape"}, base models and prompt optimization methods produce either repetitive or non-fantastical nature scenes, demonstrating both overly-similar outputs and poor text alignment. In contrast, our method generates diverse outputs, spanning mystical forests,  and supernatural landscapes, that both are visually diverse and better align with the prompt's intent.  Additionally, our method demonstrates subject-level diversity: lantern example demonstrates varied lantern types, weather conditions and locations, while other methods show limited diversity in these areas and do not align with the given prompt correctly. Furthermore, we can see diverse human subjects across age, gender, and ethnicity in kayaking scenes (please refer to Fig.\ref{fig:qual} ). 

Furthermore, \texttt{DPP-GRPO} diversifies across features unique to video models (e.g., motion and camera types). Fig.~\ref{fig:motion} demonstrates varied skateboarding maneuvers  and varied camera movements, while Fig.~\ref{fig:qual} shows diverse perspectives including aerial views, panning, and close-ups across kayaking, coffee, and cat scenes. Importantly, increased diversity does not compromise semantic fidelity: all generated videos accurately depict the core subjects and actions of the input prompt. This demonstrates that \texttt{DPP-GRPO}   balances exploration and relevance  while maintaining user intent.

\begin{table*}[t!]
\centering
\small
\setlength{\tabcolsep}{4pt}
\renewcommand{\arraystretch}{1.1}
\resizebox{\textwidth}{!}{
\begin{tabular}{l|ccc|c|ccccc|cccccc}
\hline
\multirow{2}{*}{\textbf{Model}} & 
\multicolumn{3}{c|}{\textbf{Diversity}} &
\multicolumn{1}{c|}{} &
\multicolumn{5}{c|}{\textbf{VideoScore}} &
\multicolumn{6}{c}{\textbf{VBench}} \\
 & \textsc{TCE} & \textsc{TIE} & \textsc{VENDI} 
 & \textsc{CLIP} 
 & \textsc{VQ} & \textsc{TC} & \textsc{DD} & \textsc{TVA} & \textsc{FC}
 & \textsc{SC} & \textsc{BC} & \textsc{AQ} & \textsc{IQ} & \textsc{MS} & \textsc{DD} \\
\hline
\multicolumn{16}{l}{\textbf{Wan2.1}} \\
\hline
Original prompts &19.76 & \underline{39.7} & \underline{9.2} & 0.28 & \underline{2.83} & \underline{2.76} & \underline{3.00} & \underline{2.81} & \underline{2.59} & 0.973 & 0.961 & 0.611 & 0.692 & \underline{0.991} & \textbf{0.392} \\
Promptist & \underline{19.79} & 37.58 & 8.26 & 0.29 & 2.81 & 2.71 & 2.96 & 2.78 & 2.54 & 0.975 & \underline{0.966} & 0.621 & 0.691 & \textbf{0.992} & 0.373 \\
GPT-5 & 16.71 & 29.27 & 8.15 &  0.305 &2.69 & 2.56 & 2.81 & 2.62 & 2.35 & 0.971 & \textbf{0.973} & 0.634 & 0.691 & \textbf{0.992} & 0.385 \\
Prompt-A-Video & 15.72 & 24.10 & 8.45 & \underline{0.304} & 2.74 & 2.59 & 2.95 &  2.78 & 2.39 & \textbf{0.977} & \underline{0.966} & \underline{0.652} & \underline{0.703} & 0.990 & 0.369 \\
\textbf{Ours} & \textbf{31.95} & \textbf{49.09} & \textbf{11.29} & \textbf{0.311} & \textbf{3.37} & \textbf{3.32} & \textbf{3.55} & \textbf{3.27} & \textbf{3.22} & \underline{0.976} & \underline{0.966} & \textbf{0.655} & \textbf{0.722} &\textbf{ 0.992} & \underline{0.387} \\
\hline
\multicolumn{16}{l}{\textbf{CogVideoX}} \\
\hline
Original prompts & 22.21 & 40.33 & 8.10 & 0.292 & 3.59 & 3.52 &\underline{3.73} & \underline{3.37} & 3.45 & 0.961 & 0.961 & 0.528 & 0.626 & \underline{0.990} & \textbf{0.510} \\
Promptist & \underline{23.45} &\underline{41.67} & 8.39 & 0.294 & \underline{3.60} & \underline{3.54} & 3.70 & 3.34 & \underline{3.48} & 0.954 & 0.955 & 0.524 & 0.591& \underline{0.990} & 0.493 \\
GPT-5 & 17.38 & 24.80 & \underline{8.84} & 0.296 & 2.76 & 2.64 & 2.87 & 2.50 & 2.53 & \textbf{0.964} & 0.961 & 0.561 & 0.646 & 0.985 & 0.295 \\
Prompt-A-Video  & 17.25 & 25.78 & 7.62 & \underline{0.300} & 2.86 & 2.93 & 2.70 & 2.72 & 2.76 & \underline{0.963} & \underline{0.971} & \textbf{0.651} & \textbf{0.734} & 0.987 & 0.433 \\

\textbf{Ours} & \textbf{27.59} & \textbf{45.46} & \textbf{10.30} & \textbf{0.310} & \textbf{4.08} & \textbf{4.14 }& \textbf{3.78} & \textbf{3.42} & \textbf{4.05} & \textbf{0.964} & \textbf{0.979} & \underline{0.635} & \underline{0.717} & \textbf{0.991} & \underline{0.495} \\
\hline

\end{tabular}
}
\vspace{-1em}
\caption{Quantitative comparison of our framework with baseline T2V models under two model families (\textbf{Wan2.1} and \textbf{CogVideoX}).}
\label{tab:wan_cog_table}
\end{table*}
\noindent \textbf{Quantitative Experiments} We evaluate our method against 200 prompts sampled from the VBench \cite{huang2024vbench} dataset, with 20 videos generated per prompt, resulting in 4,000 videos per method. Since there are currently no works directly targeting diverse video generation, we include several representative baselines that enhance generation quality in both image and video domains  \cite{hao2023optimizing, ji2024prompt} as well as GPT-5 \cite{achiam2023gpt}.  We assess diversity, alignment, and visual quality across a wide range of standard metrics. For diversity, we employ three complementary measures: Truncated CLIP Entropy (TCE) \cite{ibarrola2024measuring} captures 
semantic variation across samples, Truncated Inception Entropy (TIE)~\cite{ibarrola2024measuring} measures perceptual diversity in visual attributes (texture, color, lighting), and VENDI~\cite{friedman2022vendi} 
quantifies distributional diversity via eigenvalue entropy. Embeddings for TCE/TIE/VENDI are computed by sampling 8 uniformly spaced frames per video where metrics are calculated on  individual frames and averaged.
For generation quality, we use VideoScore~\cite{he2024videoscore}   and VBench~\cite{huang2024vbench} metrics.

\noindent \textbf{Video Diversity and Quality.}
Our method substantially outperforms all diversity baselines across both models (see Table \ref{tab:wan_cog_table}). On Wan2.1, we compare \texttt{DPP-GRPO} with alignment methods, Promptist \cite{li2024promptist} and Prompt-A-Video \cite{ji2025prompt} as well as GPT-5 \cite{achiam2023gpt}. we achieve substantial improvement on diversity metrics (TCE, TIE, and VENDI) over all compared methods. Similarly, on CogVideoX, diversity metrics demonstrate a significant gain in our method.
Beyond diversity metrics, our approach yields competitive  performance on \textit{VideoScore} and \textit{VBench}, indicating that increased diversity does not come at the cost of perceptual quality or fidelity. VBench results show that semantic consistency and appearance quality remain stable or slightly improved. Similar trends appear for CogVideoX, where our method achieves the strongest overall VideoScore while maintaining high semantic consistency. These results demonstrate that diversity-oriented optimization can be achieved without degrading visual or temporal coherence.

\noindent \textbf{Text Alignment.}  \texttt{DPP-GRPO} improves CLIP alignment alongside diversity (Table~\ref{tab:wan_cog_table}). CLIP scores in both CogVideoX and Wan2.1 improves compared to base models. VideoScore's TVA metric similarly improves as a result of applying our method.%

\begin{figure}%
\centering
\includegraphics[width=.48\textwidth]{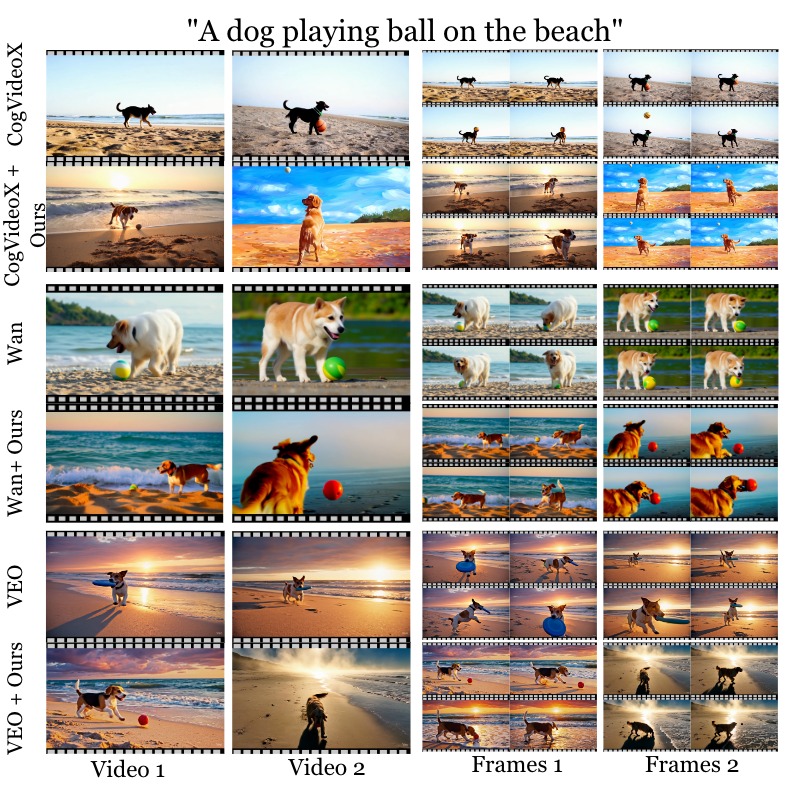}
    \vspace{-1em} 
    \caption{Visual comparison of T2V generations from various base models, with and without our method. For each baseline (CogVideoX, Wan, VEO), we show two generated videos and their representative frames. Our method enhances the diversity and quality of the generated videos across integrated models.
    }
    \vspace{-0.5em}
\label{fig:ablation-backbone}
\vspace{-0.9em}
\end{figure}

\begin{table}[t]
\centering
\vspace{-2mm}
\begin{tabular}{@{}lrc@{}}
\toprule
\textbf{Method} & \multicolumn{1}{c}{\textbf{Inference (s)}} & \textbf{Overhead} \\
\midrule
SDXL (Base) & $42.2$ & -- \\
SPARKE & $+5.25$ & +12.4\% \\
\midrule
Wan2.1 (Base) & $85.48$ & -- \\
Promptist & $+0.51$ & \textbf{+0.60\%} \\
Prompt-A-Video & $+11.02$ & +12\% \\
GPT-5 & $+22.42$ & +26\% \\
Ours & $+0.58$ & \underline{+0.67\%} \\
\bottomrule
\end{tabular}
\vspace{-0.5em}
\caption{Computational efficiency comparison. Time per video in seconds. Our method achieves minimal overhead (+0.67\%).}
\label{tab:timing}
\vspace{-0.5em}
\end{table}

\noindent \textbf{Computational Efficiency.} 
We measure average inference time per video across methods in Table~\ref{tab:timing} where we report inference time and overhead relative to base models: Wan2.1 (85.48s baseline) for video 
methods and SDXL~\cite{podell2023sdxl} (42.2s baseline) for image methods. Our method achieves 0.58s inference time with only 0.67\% overhead, second only to Promptist~\cite{li2024promptist} (0.51s, 0.60\% overhead).  Test-time optimization methods incur significantly higher costs: SPARKE~\cite{jalali2025sparke}, an image-based method built on SDXL, adds 12\% overhead per image due to iterative optimization. API-based methods also suffer from high latency: GPT-5 requires 22.42s, while Prompt-A-Video~\cite{ji2025prompt}, 
which uses Llama-3-Instruct models (8B-70B parameters), takes 11.02s. These results confirm that our approach achieves near-baseline efficiency.

\begin{table}[t]
\centering
\vspace{-1.5mm}
\begin{tabular}{@{}lcc@{}}
\toprule
\textbf{Method} & \textbf{Diversity} & \textbf{Alignment} \\
\midrule
CogVideoX        & 2.55 & 3.75 \\
Prompt-A-Video   & 2.70 & 3.83 \\
Promptist        & 3.02 & 3.97 \\
Wan2.1           & 3.10 & 3.80 \\
\textbf{Ours}    & \textbf{4.07} & \textbf{4.28} \\
\bottomrule
\end{tabular}
\vspace{-0.7em}
\caption{Our method achieves the highest ratings for both diversity and alignment in human evaluation.}
\label{tab:user_study}
\vspace{-1.8em}
\end{table}
\noindent \textbf{User Study.} We ran a randomized user study to evaluate perceived set-level diversity and prompt alignment with participants recruited from Prolific.com. For each of 10 prompts, participants viewed one set per method composed of 4 clips, with method labels hidden and both prompt and video orders counterbalanced to mitigate position and fatigue effects.  After viewing each set, participants rated Diversity and Alignment on 5-point Likert scales. \texttt{DPP-GRPO} received significantly higher Diversity scores than all baselines, indicating broader set-level variety. It also achieved higher Alignment on average, while competing methods performed closely, as expected given their strong prompt adherence. \textit{Please refer to SM for more details.}

\subsection{Ablation Studies.} We conduct a series of ablations to analyze the contribution of different components of \texttt{DPP-GRPO} and validate the robustness of our design choices.

\begin{table*}[t!]
\centering
\small
\setlength{\tabcolsep}{4pt}
\renewcommand{\arraystretch}{1.1}
\resizebox{\textwidth}{!}{
\begin{tabular}{l|ccc|c|ccccc|cccccc}
\hline
\multirow{2}{*}{\textbf{Model}} & 
\multicolumn{3}{c|}{\textbf{Diversity}} &
\multicolumn{1}{c|}{} &
\multicolumn{5}{c|}{\textbf{VideoScore}} &
\multicolumn{6}{c}{\textbf{VBench}} \\
 & \textsc{TCE} & \textsc{TIE} & \textsc{VENDI} 
 & \textsc{CLIP} 
 & \textsc{VQ} & \textsc{TC} & \textsc{DD} & \textsc{TVA} & \textsc{FC}
 & \textsc{SC} & \textsc{BC} & \textsc{AQ} & \textsc{IQ} & \textsc{MS} & \textsc{DD} \\
\hline
\multicolumn{16}{l}{\textbf{Wan2.1}} \\
\hline
Ours (SFT) & 14.05 & 17.13 & 11.05 & 0.251 & 2.83 & 2.72 & 2.99 & 2.79 & 2.56 & 0.962 & 0.960 & 0.623 & 0.700 & 0.989 &0.266 \\
Ours (only relevance) & 20.06 & 35.66 & 8.87 & 0.283 & 2.60  & 2.55 & 2.85 & 2.61 & 2.45 & 0.971 & 0.964 & 0.596 & 0.702 & 0.981 & 0.247 \\
Ours (only DPP) & 27.05 & 45.21 & \textbf{11.72} & 0.255 & 2.74 & 2.58 &2.86 & 2.92 & 2.15 & 0.961 & 0.972 & 0.597 & 0.705 & 0.984 & 0.247 \\
\textbf{Ours (full)} & \textbf{31.95} & \textbf{49.09} & 11.29 & \textbf{0.311} & \textbf{3.37} & \textbf{3.32} & \textbf{3.55} & \textbf{3.27} & \textbf{3.22} & \textbf{0.976} & \textbf{0.966} & \textbf{0.655} & \textbf{0.722} &\textbf{ 0.992} & \textbf{0.387} \\
\hline
\end{tabular}
}
\vspace{-1em}
\caption{Ablation studies conducted with Wan backbone over different reward terms, supervised-finetuned version, and our full complete model}
\label{tab:ablation-tab}
\end{table*}
\noindent \textbf{Ablation on different backbones} To evaluate model-agnostic performance, we ablate across two state-of-the-art text-to-video backbones: Wan2.1 and CogVideoX. As shown in Table~\ref{tab:ablation-tab}, \texttt{DPP-GRPO} consistently improves both semantic and visual diversity metrics across architectures, demonstrating that our framework generalizes well regardless of the underlying diffusion model. The improvements are particularly notable in metrics such as TCE, TIE, and VENDI, confirming that our diversity modeling is not tied to any specific video model. Moreover, Fig. \ref{fig:ablation-backbone} presents a side-by-side comparison applying our method to Wan, CogVideoX, and Veo 3. As can be seen from the visuals, the results demonstrate diverse outputs across various backbones.
\begin{figure}%
\centering
\includegraphics[width=0.35\textwidth]{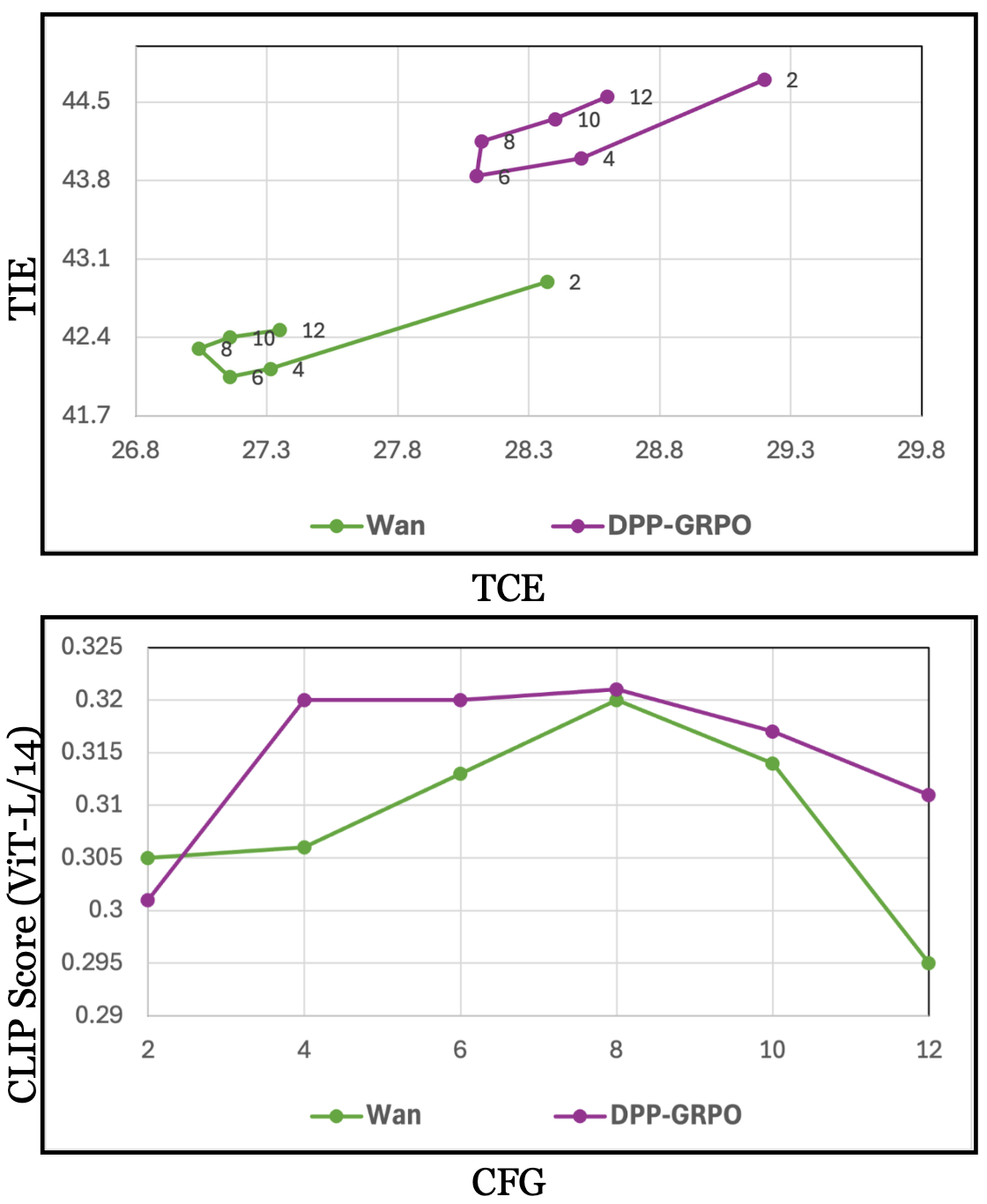}
    \vspace{-0.75em} 
    \caption{
\textbf{CFG ablation.} 
Diversity (TCE/TIE) and fidelity (CLIP) across different CFG values for Wan and our DPP--GRPO model.
}

\label{fig:cfg-ablation}
\vspace{-2em}
\end{figure}
\noindent \textbf{Ablation on reward terms} We further analyze the impact of individual reward components by isolating the diversity and relevance terms in our composite objective in Eq.~\ref{eq:combined_reward}. Using only the relevance term yields higher CLIP alignment scores, as expected, yet significantly reduces diversity, often collapsing to repetitive outputs similar to base models. Conversely, using only DPP-based diversity term increases variation across motion and scene attributes but leads to a modest decline in text–video consistency. Our full model, which balances both terms, achieves the best balance between semantic fidelity and diversity, improving across both dimensions simultaneously. This confirms that the dual-term reward formulation is essential for diverse video generation.
\begin{figure*}%
\centering
\includegraphics[width=0.8\textwidth]{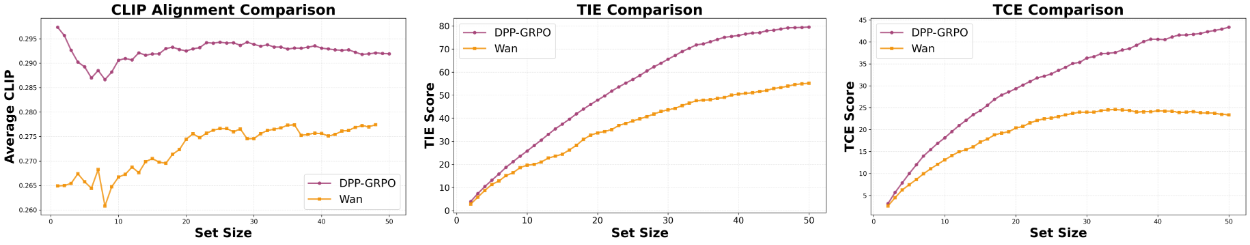}
    \vspace{-1em}
    \caption{Impact of generation set size on diversity and alignment. 
        The plots compare our \texttt{DPP-GRPO} method against the Wan baseline on 
        (Left) Average CLIP alignment, (Middle) semantic diversity (TCE), 
        and (Right) perceptual diversity (TIE). 
        \texttt{DPP-GRPO} consistently outperforms the baseline in both alignment 
        and diversity as the set grows.}
    \vspace{-1em}
\label{fig:ablation}
\end{figure*}

\begin{figure}%
\centering
\includegraphics[width=0.4\textwidth]{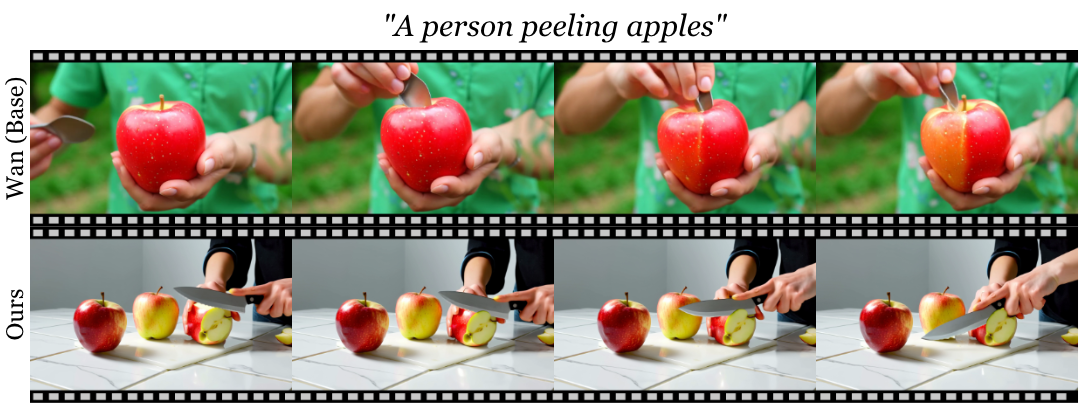}
    \vspace{-0.75em} 
    \caption{An example failure case of our method where the temporal video quality depends on the base model's ability.}\vspace{-1em}
\label{fig:failure-case}
\end{figure}
\noindent \textbf{Ablation on generation set size}
We analyze the impact of the generation set size during inference, with results shown in Fig. \ref{fig:ablation}. Figure plots the running TIE, TCE, and average CLIP scores as the set of generated videos grows, comparing \texttt{DPP-GRPO} to the Wan baseline. The diversity plots reveal two key findings: Our method consistently achieves higher semantic (TCE) and perceptual (TIE) diversity scores than the baseline at every set size. While both methods exhibit diminishing returns, where the marginal diversity gain from each new video decreases as the set expands, the baseline model's diversity plateaus much earlier. In contrast, \texttt{DPP-GRPO} continues to discover varied outputs, achieving significantly higher overall diversity.
Diversity gain do not compromise prompt alignment. The CLIP Alignment plot shows that \texttt{DPP-GRPO} maintains a stable average CLIP score compared to the Wan baseline. As a result, our method successfully expands diversity while simultaneously maintaining its semantic fidelity.

\noindent\paragraph{Effect of classifier-free guidance} 
Figure~\ref{fig:cfg-ablation} presents an ablation study measuring the effect of
classifier-free guidance (CFG) values on both diversity and fidelity for the 
baseline Wan model and our \texttt{DPP-GRPO} method. We sweep the CFG scale over 
$\{2,4,6,8,10,12\}$ and report (1) the joint behavior of TCE and TIE, and (2)
CLIP alignment scores.
The upper plot in Figure~\ref{fig:cfg-ablation} shows that our method consistently achieves higher TCE and TIE across all CFG values. While both models exhibit non-monotonic trajectories as CFG increases, the \texttt{DPP-GRPO} frontier shifts 
upward in the TCE-TIE plane, demonstrating that our objective is robust to abrupt CFG adjustments. This indicates that our method 
remains effective under a wide range of decoding hyperparameters and does not depend on carefully tuned CFG settings.
The lower plot in Figure~\ref{fig:cfg-ablation} reports CLIP similarity scores. Both models exhibit the expected behavior in which CLIP 
scores peak at moderate guidance scales ($\text{CFG}\approx 6$-$8$). Importantly,  our method maintains fidelity comparable to the baseline across all settings, and achieves the highest CLIP score at $\text{CFG}=6$.

\section{Discussion}

\noindent \textbf{Limitations and Broader Impact} Our approach inherits base model limitations in complex temporal dynamics as temporal dynamics are governed by the model's temporal attention values rather than prior conditioning \cite{shaulov2025flowmo}. Figure~\ref{fig:failure-case} shows when the base  model struggles with fine-grained motions in complex actions (e.g., peeling actions), our method also struggles in overcoming this constraint,
Nevertheless, our work aims to make T2V generation diverse by design, reducing the need for compute-intensive trial-and-error.  This can reduce operational cost and improve accessibility for creators or  small teams that lack large compute budgets.

\noindent \textbf{Conclusion} We introduced \texttt{DPP-GRPO}, a set-level policy optimization framework that treats diversity as an explicit optimization objective for text-to-video generation.  Our method combines group-relative policy optimization with a DPP-based diminishing-returns term, encouraging complementary coverage across cinematographic and semantic factors while preserving prompt fidelity. Our approach is model-agnostic and plug-and-play; operating in prompt space to work with both open models (Wan, CogVideoX) and black-box systems (e.g., Veo) and directly yields diverse sets without post-hoc re-ranking or cherry-picking. Experiments on several benchmarks,  human studies, and qualitative comparisons demonstrate consistent gains in set-level diversity with equal or better alignment and perceptual quality.

\clearpage
  {
      \small
      \bibliographystyle{ieeenat_fullname}
     \bibliography{main}
 }
\clearpage
\setcounter{page}{1}
\maketitlesupplementary

\section{Additional Qualitative Examples}
Please refer to the end of supplementary webpage (.html file) or the supplementary folder for all the videos in our main paper, as well as additional  videos.

\begin{table}[h]
\centering

\begin{tabular}{cccc}
\toprule
\textbf{Set Size} & \textbf{TCE} & \textbf{TIE} & \textbf{CLIP} \\
\midrule
2  & 15.131 & 22.939 & 0.308 \\
5  & 16.570 & 23.936 & 0.308 \\
8  & 16.959 & 24.464 & 0.306 \\
10 & 15.546 & 23.040 & 0.306 \\
\bottomrule

\end{tabular}
\caption{Ablation on reference set size $\lvert \mathcal{R}_q \rvert$. 
Small multi-reference sets (5--8 examples) yield the best diversity, while larger sets show diminishing returns.}
\label{tab:refsize}
\end{table}
\begin{table}[h]
\centering

\begin{tabular}{cccc}
\toprule
\textbf{($\lambda_{\text{div}}$, $\lambda_{\text{rel}}$)} & \textbf{TCE} & \textbf{TIE} & \textbf{CLIP} \\
\midrule
(0.9, 0.1) & 16.910 & 24.256 & 0.285 \\
(0.5, 0.5) & 16.709 &  23.945 & 0.302\\
(0.1, 0.9) & 16.002 & 23.735& 0.305 \\

\bottomrule
\end{tabular}
\caption{Ablation on reward weights $(\lambda_{\text{div}}, \lambda_{\text{rel}})$.
We vary the balance between diversity and relevance during training.}
\label{tab:lambda-ablation}
\end{table}
\section{Video Prompts} 
Due to space limitations, we could not include the full prompts in the main paper. However, we provide the complete prompts for several videos generated by our method as well as  and system prompt used in tables \ref{tab:boat}, \ref{tab:cat}, \ref{tab:coffee}, \ref{tab:dog}, \ref{tab:fantasy}, \ref{tab:giraffe}, \ref{tab:kayak}, \ref{tab:skateboarder}, and \ref{tab:system_prompt_small}.

\section{User Study Details} 

\begin{figure*}%
\centering
\includegraphics[width=\textwidth]{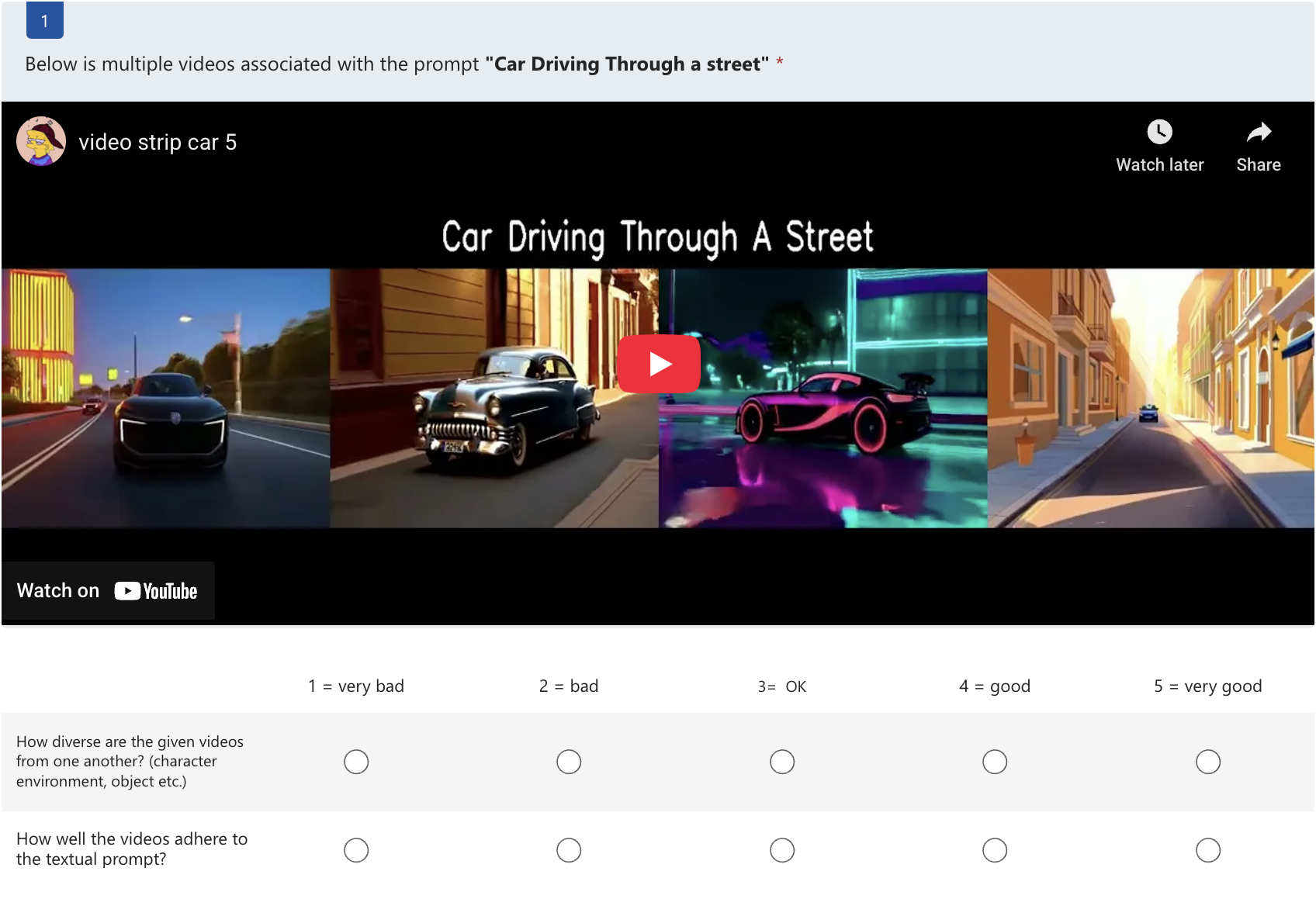}
    \vspace{-0.75em} 
    \caption{A screenshot of our user study where 4 videos from a given method are shown to the users.}\vspace{-1em}
\label{fig:user_detail}
\end{figure*}

We provide a screenshot of our user study in Fig. \ref{fig:user_detail}. Users are shown 4 videos generated by a given method, and asked to rate the diversity and text alignment on a Likert scale 1-5.

\section{Additional Ablations}

\subsection{Reference Set Size during Training}

We study how the size of the reference set $\lvert \mathcal{R}_q \rvert$ influences training dynamics and diversity. 
Table~\ref{tab:refsize} shows that using a small multi-reference set (5-8 examples) yields the higher improvements in TCE and TIE, confirming that exposing the policy to several video-grounded modes produces a more reliable marginal-gain signal. 
Performance degrades when the set becomes too large ($\lvert \mathcal{R}_q \rvert=10$), consistent with the diminishing-returns property of the DPP log-determinant and the increased variance of similarity estimates in larger matrices. 
CLIP alignment remains stable across settings, with only a mild drop for larger sets. 
Overall, a modest reference set (5-8 samples) provides the best balance between diversity gain and semantic stability.
\subsection{Ablation on $\lambda$ hyperparameters}
Table~\ref{tab:lambda-ablation} shows how varying the reward weights $(\lambda_{\text{}}, \lambda_{\text{div}})$ affects performance. Increasing the diversity weight ($\lambda_{\text{div}} = 0.9$) yields the highest TIE/TCE, indicating stronger diversity, while moderately CLIP alignment. Conversely, prioritizing relevance ($\lambda_{\text{rel}} = 0.9$) improves CLIP but reduces diversity. The balanced setting $(0.5, 0.5)$ provides a middle ground across all metrics. Overall, the ablation highlights the expected trade-off: higher diversity weight improves motion and variation, while higher relevance weight preserves semantic fidelity.

\section{Dataset Generation}
Our dataset is constructed in two stages. First, we extend the VBench prompt categories by using chain-of-thought prompting to generate approximately 350 prompts for each of the 7 VBench categories. We additionally include the original VBench prompts as part of our evaluation set. This ensures that our prompt space remains grounded in a widely used video-generation benchmark.
In the second stage, each curated base prompt is expanded into 10 diverse variants using an iterative, two-agent reasoning framework. An architect agent proposes candidate expansions (system prompt shown in Table~\ref{tab:system_prompt_small}), and a critic agent evaluates the videos produced from these expansions using established video-level metrics: TCE/TIE for temporal diversity, CLIP for semantic alignment, and VideoScore for perceptual quality. Only expansions that satisfy diversity and alignment criteria are retained. This procedure ensures that the final dataset is not only textually diverse but also grounded in video-level behavior.

\begin{table*}[t]
\centering
\caption{System Prompt}
\label{tab:system_prompt_small}

\begin{tcolorbox}[
    title=\textbf{System Prompt Content}, 
    colframe=blue!60!black, 
    colback=white,           
    coltitle=white,          
    fonttitle=\bfseries, 
    width=\textwidth,        
    arc=2mm,                 
    boxrule=1.5pt            
]
\ttfamily \scriptsize 
\hspace*{4mm} You are an expert at prompt expansion. Given a base prompt, expand it into at most four sentences.\\
\hspace*{8mm} \textbf{1. Preserve the main object(s), number, and action (no semantic drift).}\\
\hspace*{8mm} 2. Expand along dimensions: environment, object attributes, subject variation, time/season, perspective, narrative/action, character appearance, or cultural context.\\
\hspace*{8mm} 4. Use English only.\\
\hspace*{8mm} \textbf{Example}:\\
\hspace*{8mm} \textbf{Base prompt}: a cat sitting on a windowsill.\\
\hspace*{8mm} \textbf{Good expansion}: "A fluffy Siamese cat with bright blue eyes sits on a wooden windowsill, watching the rainy evening street in a warm painterly style."\\
\hspace*{8mm} \textbf{Bad expansion}: "A group of cats running through a sunny garden, chasing butterflies." (changes number, action, setting).\\
\hspace*{4mm} You will be given one user query.\\
\hspace*{4mm}\textbf{ Your task}: write a new expansion that is diverse from the references while still preserving meaning.

\end{tcolorbox}
\end{table*}
\begin{table*}[t]
\centering
\caption{Video Prompts}
\label{tab:water_lily}

\begin{tcolorbox}[
    title=\textbf{Water Lily rests on a calm pond}, 
    colframe=blue!60!black, 
    colback=white,           
    coltitle=white,          
    fonttitle=\bfseries, 
    width=\textwidth,        
    arc=2mm,                 
    boxrule=1.5pt            
]
\ttfamily \scriptsize 
\hspace*{4mm} \textbf{Video 1}: "A minimalist depiction of a white water lily resting at the edge of a calm pond, bold outlines, restrained greens, and a crisp lines."\\
\hspace*{4mm} \textbf{Video 2}: "A soft watercolor scene showing a water lily resting on a calm pond, the bloom painted in gentle pink tones with a soft wash and a gentle backdrop of water reflections."\\
\hspace*{4mm} \textbf{Video 3}:  "A top-down, aerial view of a water lily resting on a calm pond, the outline tracing a perfect circle while the pool's mirror-like surface reflects the scene in photorealistic detail."\\
\hspace*{4mm} \textbf{Video 4}:  "A minimalist vector style renders a water lily resting at the edge of a calm pond, the lotus represented as a clean silhouette against a calm, mirror-like surface with subtle radial symmetry."\\

\end{tcolorbox}
\end{table*}

\begin{table*}[t]
\centering
\caption{Video Prompts}
\label{tab:cat}

\begin{tcolorbox}[
    title=\textbf{A cat eating food out of a bowl on a sidewalk}, 
    colframe=blue!60!black, 
    colback=white,           
    coltitle=white,          
    fonttitle=\bfseries, 
    width=\textwidth,        
    arc=2mm,                 
    boxrule=1.5pt            
]
\ttfamily \scriptsize 
\hspace*{4mm} \textbf{Video 1}: "A shot of a cat on a sunlit sidewalk eating from a shallow ceramic bowl, food a bright color, street life and people blurred in the background."
\\
\hspace*{4mm} \textbf{Video 2}: "In a street cafe atmosphere, a Siamese cat eats from a ceramic bowl on a sunlit sidewalk, the bowl catching the glow and the cat's fur shimmering with the passing traffic in soft watercolor."\\
\hspace*{4mm} \textbf{Video 3}:  "In a sunlit courtyard, a fluffy Persian cat sits on a bright cobblestone, eating kibble from a ceramic bowl while the sun casts a warm glow across the scene."\\
\hspace*{4mm} \textbf{Video 4}:  "From a ground-level perspective, a cat in a bold patterned jacket eats from a bright ceramic bowl on a sunlit city sidewalk, the bowl's rim catching a glint of sun while the background bustles with activity"\\

\end{tcolorbox}
\end{table*}

\begin{table*}[t]
\centering
\caption{Video Prompts}
\label{tab:boat}

\begin{tcolorbox}[
    title=\textbf{A boat gliding across a lake at twilight}, 
    colframe=blue!60!black, 
    colback=white,           
    coltitle=white,          
    fonttitle=\bfseries, 
    width=\textwidth,        
    arc=2mm,                 
    boxrule=1.5pt            
]
\ttfamily \scriptsize 
\hspace*{4mm} \textbf{Video 1}: "A moody urban scene: a sleek glass-bottomed ferry glides across a glittering lake at twilight, the city skyline glowing behind and silhouettes etched in the glass, captured in digital painting with a dramatic color gradient."
\\
\hspace*{4mm} \textbf{Video 2}: "A cinematic time-lapse across a lake at twilight showing a boat gliding across the water, the sky's changing light refracting on the surface and the water reflecting a pale, dreamlike color."
\\
\hspace*{4mm} \textbf{Video 3}:  "Across a lake, a boat glides at twilight; the camera tracks the craft from a low-angle perspective on a grassy bank, the water shifting from silver to pink and the horizon turning a deep indigo."\\
\hspace*{4mm} \textbf{Video 4}: "A hyperreal CGI rendering of a sleek boat gliding across a glassy lake at twilight, reflective water, starry skies, and cool blue-green hues to capture the moment."\\

\end{tcolorbox}
\end{table*}

\begin{table*}[t]
\centering
\caption{Video Prompts}
\label{tab:giraffe}

\begin{tcolorbox}[
    title=\textbf{A giraffe bending to sip water from a sunlit savanna pool}, 
    colframe=blue!60!black, 
    colback=white,           
    coltitle=white,          
    fonttitle=\bfseries, 
    width=\textwidth,        
    arc=2mm,                 
    boxrule=1.5pt            
]
\ttfamily \scriptsize 
\hspace*{4mm} \textbf{Video 1}: "From a low-angle view, a giraffe bends to sip water from a sunlit savannah pool, long eyelashes brushing the surface and the sunlit grasses shimmering in the background"\\
\hspace*{4mm} \textbf{Video 2}: "A painterly color-graded shot with warm sunset hues shows a giraffe bending to sip water from a sunlit savanna pool, distant grasses glow and a subtle horizon line guides the eye.",\\
\hspace*{4mm} \textbf{Video 3}:  "A painterly color-graded frame turning the sunlit savanna into amber tones, a giraffe bending to sip water from a sunlit savanna pool, rendered in digital gouache with thick brushstrokes.",\\
\hspace*{4mm} \textbf{Video 4}: "A photorealistic close-up of a giraffe bending to sip water from a sunlit savanna pool, as the sunbeams split the grasses and the water shimmers with color."\\

\end{tcolorbox}
\end{table*}

\begin{table*}[t]
\centering
\caption{Video Prompts}
\label{tab:skateboarder}

\begin{tcolorbox}[
    title=\textbf{A skateboarder performs jumps}, 
    colframe=blue!60!black, 
    colback=white,           
    coltitle=white,          
    fonttitle=\bfseries, 
    width=\textwidth,        
    arc=2mm,                 
    boxrule=1.5pt            
]
\ttfamily \scriptsize 
\hspace*{4mm} \textbf{Video 1}: " "On a sunlit park plaza, a teenage Black girl with vibrant hair performs a set of jumps and grinds on a street-style board, ground reflects light."\\
\hspace*{4mm} \textbf{Video 2}: "On a concrete skatepark plaza at sunset, a Black teenage skateboarder performs a series of jumps from a deck, the crowd watching as the skateboarder leaps with confidence, captured in cinematic shot."\\

\end{tcolorbox}
\end{table*}

\begin{table*}[t]
\centering
\caption{Video Prompts}
\label{tab:fantasy}

\begin{tcolorbox}[
    title=\textbf{A fantasy landscape}, 
    colframe=blue!60!black, 
    colback=white,           
    coltitle=white,          
    fonttitle=\bfseries, 
    width=\textwidth,        
    arc=2mm,                 
    boxrule=1.5pt            
]
\ttfamily \scriptsize 
\hspace*{4mm} \textbf{Video 1}:"A panorama-style shot of a forest glade at twilight, holographic flowers and glow moss lighting the scene as a breeze moves the leaves."
\\
\hspace*{4mm} \textbf{Video 2}: "A surreal yet coherent dreamlike landscape where frost-draped trees shimmer under aurora, the moonlight catching tiny frost crystals, and a faint, otherworldly air pervades the scene."\\
\hspace*{4mm} \textbf{Video 3}:  "In a world where magic meets science, a sun-dappled desert rises around a mysterious castle; crystals glow with luminosity as water droplets create pools across the scene."\\
\hspace*{4mm} \textbf{Video 4}: "A dawn panorama across a fantastical valley with a river, a shimmering dragonfly-like wing across the sky, and the air thick with mist, rendered in pastel watercolor with a warm glow."\\

\end{tcolorbox}
\end{table*}

\begin{table*}[t]
\centering
\caption{Video Prompts}
\label{tab:coffee}

\begin{tcolorbox}[
    title=\textbf{A coffee cup sitting on a wooden table}, 
    colframe=blue!60!black, 
    colback=white,           
    coltitle=white,          
    fonttitle=\bfseries, 
    width=\textwidth,        
    arc=2mm,                 
    boxrule=1.5pt            
]
\ttfamily \scriptsize 
\hspace*{4mm} \textbf{Video 1}:"A cinematic top-down view in a modern kitchen shows a porcelain coffee cup on a wooden table, a single grain visible in the wood and a single drop of condensation forming on the rim."
\\
\hspace*{4mm} \textbf{Video 2}: "An artisanal coffee mug with a rustic ceramic rim rests on a wooden table, the scene rendered in warm watercolor with soft brushstrokes and a delicate steam curl above the mug"\\
\hspace*{4mm} \textbf{Video 3}: "A studio macro on a glass coffee cup on a wooden table, the cup resting in the center while the wooden grain and the steam in the air form a high-contrast backdrop.",
\\
\hspace*{4mm} \textbf{Video 4}:  "A vintage-inspired setting shows a glass coffee cup on a wooden table, sunbeams filtering through lace curtains and dust motes dancing in the air as the cup glows softly."\\

\end{tcolorbox}
\end{table*}

\begin{table*}[t]
\centering
\caption{Video Prompts}
\label{tab:kayak}

\begin{tcolorbox}[
    title=\textbf{A person kayaking or canoeing}, 
    colframe=blue!60!black, 
    colback=white,           
    coltitle=white,          
    fonttitle=\bfseries, 
    width=\textwidth,        
    arc=2mm,                 
    boxrule=1.5pt            
]
\ttfamily \scriptsize 
\hspace*{4mm} \textbf{Video 1}:"A South Asian woman in a wetsuit paddles a bright pink kayak across a tranquil river, sunlight turning the water to copper and casting a warm halo around the boat."
\\
\hspace*{4mm} \textbf{Video 2}:  "A dawn scene where a person paddles a kayak toward a misty marsh; soft washes of pink and teal, with a shallow, warm glow on the water."\\
\hspace*{4mm} \textbf{Video 3}: "A sunlit meadow scene where a person in a neon-green paddling jacket is canoeing along a quiet river, wildflowers blooming nearby and the sun casting a warm glow across the water."
\\
\hspace*{4mm} \textbf{Video 4}: "A person in a kayak glides along a tranquil river at dawn, sun rising and the water shimmering with golds, rendered in dreamy mood with loose brushwork and soft edges."
\\

\end{tcolorbox}
\end{table*}

\begin{table*}[t]
\centering
\caption{Video Prompts}
\label{tab:dog}

\begin{tcolorbox}[
    title=\textbf{A dog playing ball on the beach}, 
    colframe=blue!60!black, 
    colback=white,           
    coltitle=white,          
    fonttitle=\bfseries, 
    width=\textwidth,        
    arc=2mm,                 
    boxrule=1.5pt            
]
\ttfamily \scriptsize 
\hspace*{4mm} \textbf{Video 1}:In a sunset beach scene, a beagle with a bright coat and long ears romps after a ball on the sand, waves lapping and a light breeze curling the hair, watercolor mood."
\\
\hspace*{4mm} \textbf{Video 2}:  "A cinematic color-graded shot with vibrant warm hues and cool shadows of a golden retriever at foggy weather, playing with a ball on a sunlit beach, high-angle view from above."\\

\end{tcolorbox}
\end{table*}

\begin{table*}[t]
\centering
\caption{Video Prompts}
\label{tab:fox}

\begin{tcolorbox}[
    title=\textbf{A fox walking through a forest}, 
    colframe=blue!60!black, 
    colback=white,           
    coltitle=white,          
    fonttitle=\bfseries, 
    width=\textwidth,        
    arc=2mm,                 
    boxrule=1.5pt            
]
\ttfamily \scriptsize 
\hspace*{4mm} \textbf{Video 1}:"A photorealistic dawn scene of a red fox walking through a forest, warm light filtering through a canopy of birches and needles, dewy moss and a distant stream catching the early sun."
\\
\hspace*{4mm} \textbf{Video 2}:"A white fox walking through a sunlit forest, soft brushwork greens and browns; the scene is dreamlike with a slight hint of surrealism."
\\
\hspace*{4mm} \textbf{Video 3}: "A surreal concept envisions a fox walking through a dreamlike forest, trees bending to form a path and light washing the fur in soft tones."
\\
\hspace*{4mm} \textbf{Video 4}: "A hyper-realistic CGI depiction of a fox walking through a forest at dawn, thick fur rendered with photoreal texture, golden hour light highlighting the forest floor."
\\

\end{tcolorbox}
\end{table*}

\begin{table*}[t]
\centering
\caption{Video Prompts}
\label{tab:cyvlist}

\begin{tcolorbox}[
    title=\textbf{A cyclist riding along a lakeside trail}, 
    colframe=blue!60!black, 
    colback=white,           
    coltitle=white,          
    fonttitle=\bfseries, 
    width=\textwidth,        
    arc=2mm,                 
    boxrule=1.5pt            
]
\ttfamily \scriptsize 
\hspace*{4mm} \textbf{Video 1}:"A sunset sequence shows a cyclist riding along a lakeside trail, the sun casting warm light on the rider and the water reflecting hues of pink."
\\
\hspace*{4mm} \textbf{Video 2}:"A vintage style rendering of a cyclist riding along a lakeside trail, bold shapes and warm sunset hues, retro typography for the route and destination."
\\
\hspace*{4mm} \textbf{Video 3}: "A painterly color-graded frame turning the greens to emerald and the sky to a pale cerulean, as the cyclist glides along a lakeside trail, water reflections shimmering in the frame"
\\
\hspace*{4mm} \textbf{Video 4}: "A cyclist on a lakeside trail glides toward the camera, reflections on the water shimmering and ripples spreading as a tranquil backdrop renders the scene cinematic."
\\

\end{tcolorbox}
\end{table*}

\begin{table*}[t]
\centering
\caption{Video Prompts}
\label{tab:lantern}

\begin{tcolorbox}[
    title=\textbf{A lantern swaying softly on windy night}, 
    colframe=blue!60!black, 
    colback=white,           
    coltitle=white,          
    fonttitle=\bfseries, 
    width=\textwidth,        
    arc=2mm,                 
    boxrule=1.5pt            
]
\ttfamily \scriptsize 
\hspace*{4mm} \textbf{Video 1}:"A painterly color-graded scene where the lantern shifts to warm amber, the night breeze makes the light flutter and the background blurs into a dreamlike wash."
\\
\hspace*{4mm} \textbf{Video 2}:"A painterly night scene showing a lantern swaying softly on a windy night in a quiet street, the glow diffusing through the wind and casting a warm halo across the cobblestones"
\\
\hspace*{4mm} \textbf{Video 3}: "Isometric landscape vignette: a lantern floats in a serene park at twilight; ornate, graphic lanterns define natural shapes in geometric forms; soft pastel sky overhead adds a cozy, calm mood."
\\
\hspace*{4mm} \textbf{Video 4}: "A cinematic long-shot with a lantern swaying softly on a windy night, a coastal boardwalk as the backdrop, mist rolling and the lantern creating a warm halo around the walker."
\\

\end{tcolorbox}
\end{table*}

\end{document}